\documentclass{article}

\usepackage{PRIMEarxiv}

\usepackage[utf8]{inputenc} 
\usepackage[T1]{fontenc}    
\usepackage{hyperref}       
\usepackage{url}            
\usepackage{booktabs}       
\usepackage{amsfonts}       
\usepackage{nicefrac}       
\usepackage{microtype}      
\usepackage{lipsum}
\usepackage{fancyhdr}       
\usepackage{graphicx}       
\graphicspath{{media/}}     
\usepackage{wrapfig}
\usepackage{amsmath}
\usepackage{algorithm}
\usepackage{algpseudocode}
\usepackage{subcaption}
\usepackage{array}
\newtheorem{assumption}{Assumption}

\title{Neural Expectation-Maximization and meta-models
}

\author{
  Zhuolin Fu \\
  \texttt{fzhuolin@gmail.com} \\
}

\begin{document}
\maketitle

\begin{abstract}
We interpret Transformer networks as learned approximations to a hybrid probabilistic-inference procedure that combines max-product Loopy Belief Propagation (LBP) with Expectation--Maximization (EM). In this view, a Transformer performs inference in partially observed Markov Random Fields (MRFs) whose structure and parameters are not given a priori but are inferred from data. We show that self-attention implements a differentiable approximation to max-product LBP on an input-conditioned, dynamically structured MRF, with attention weights encoding learned structural affinities. We further establish an EM correspondence at the level of depth: each layer updates latent representations in an E-step-like forward pass, while the transition $\theta^{(l)} \to \theta^{(l+1)}$ across layers acts as an implicit M-step. Building on this interpretation, we introduce \emph{meta-models}, a design paradigm in which inputs generate instance-specific parameters, unifying mechanisms such as self-attention, mixture-of-experts, and gated units within a single framework. Empirical results support the proposed perspective.

\end{abstract}


\section{Introduction}
\begin{figure}[h]
    \centering
    \begin{minipage}[b]{0.45\textwidth}
        \includegraphics[width=0.8\linewidth]{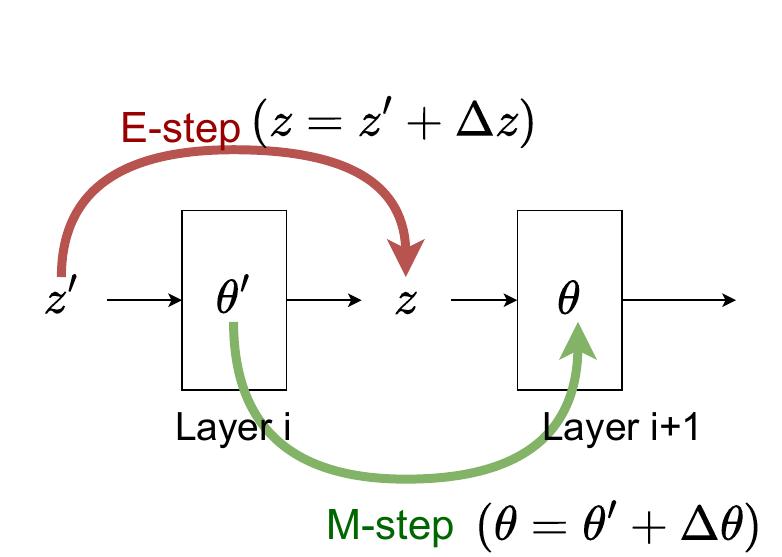}
        \caption{
        Expectation-Maximization interpretation of DNNs.
        A layer corresponds to an iteration of the EM. The forward pass corresponds to the E-step.
        The transition from parameter $\theta'$ to the next layer $\theta$ implicitly corresponds to the M-step.
        }
        \label{fig:dnn_as_em}
    \end{minipage}
    \hspace{0.08\textwidth}
    \begin{minipage}[b]{0.45\textwidth}
        \centering
        \includegraphics[width=\linewidth]{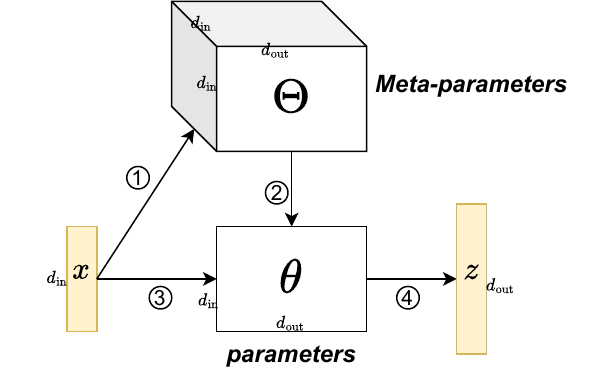}
        \caption{
        Illustration of a meta-model.
        The input vector $x$ first interacts with the 3D meta-parameter $\Theta$ and outputs the 2D parameter $\theta$.
        Then, $x$ interacts with $\theta$ and produces the output $z$.
        Meta-model is what makes a model capable of in-context learning.
        Self-attention, MoE, GLUs, etc. are members of meta-models.
        }
        \label{fig:meta-models}
    \end{minipage}
\end{figure}

Transformers have become the dominant architecture for modern sequence modeling, language understanding, and large-scale generative modeling since the introduction of self-attention~\cite{vaswani2017attention}.
Despite their empirical success, the internal computation performed by a Transformer remains difficult to characterize in algorithmic terms.
A growing body of work has therefore shifted from treating Transformers as black-box function approximators toward explaining what their layers, attention heads, and residual streams compute~\cite{elhage2021mathematical,olsson2022context}.
This paper contributes to that line of work by asking whether Transformer computation can be understood as a familiar inference procedure rather than as a collection of isolated architectural mechanisms.

Existing explanations illuminate important aspects of Transformer behavior, but they remain partial.
Studies of in-context learning show that Transformers can implement gradient-descent-like or Bayesian updates; and message-passing views relate attention to graph computation~\cite{akyurek2023learning,dai2023gpt,xie2022explanation,velickovic2018graph}.
However, these perspectives do not fully account for how self-attention, depth, latent-variable refinement, graph-structure selection, and layerwise parameter changes fit into a single probabilistic picture.

We propose such an interpretation by viewing Transformers as neuralized Expectation-Maximization.
Under this view, the underlying model is a partially observed Markov Random Field (MRF) with both unknown structure and parameters.
Self-attention approximates Loopy Belief Propagation (LBP) over a dynamically inferred graph structure, where the attention between two tokens acts as the soft connection strength within such MRF and the value vector acts as the maximizer of the aggregated message (aka. belief).
Successive layers play the role of Expectation-Maximization (EM) iterations~\cite{dempster1977maximum,yedidia2003understanding,koller2009probabilistic} for parameter learning.
Specifically, a layer of a Transformer corresponds to an iteration of the EM.
The forward pass of a layer corresponds to the E-step, where the latent variables are updated based on the last iteration.
The change from the parameters $\theta^{(l)}$ of layer $l$ to $\theta^{(l+1)}$ of the next layer implicitly corresponds to the M-step, where the parameters are updated.

This paper makes the following contributions:
\begin{itemize}
    \item We derive a correspondence between self-attention and approximate LBP on a dynamically structured MRF, where attention weights act as learned structural affinities and values approximate individual message maximizers.
    \item We interpret the depth of a Transformer as an EM-like iterative procedure in which the forward pass updates latent variables and the transition between layer parameters corresponds to parameter refinement.
    \item We use this EM perspective to motivate meta-models, a design paradigm in which the parameters $\theta$ of a forward pass are generated from the input by meta-parameters $\Theta$.
    \item We show that self-attention, Mixture-of-Experts layers, gated mechanisms, LSTM, and selective state-space models can be viewed as members or close relatives of this meta-model principle.
\end{itemize}


\section{Related Work}
\label{sec:related_work}
Our work connects three lines of research: interpretations of transformers as probabilistic inference, interpretations of transformer layers as expectation--maximization (EM) steps, the view of attention as a meta-model that generates parameters from the input.

\paragraph{Transformers and expectation--maximization.}
A line of work interprets transformer depth as iterative probabilistic fitting.
\cite{chen2025transformersunsupervised} prove that transformers can approximate the EM algorithm, together with spectral (tensor power) iterations, on Gaussian mixtures.
The first is an interpretation tied to a specific modeling choice, and the second is an expressivity result for a particular model class.
The distinction is that our analysis frames the algorithm as probabilistic latent-variable inference: the E-step updates latent variables $z$, while changes in layer parameters $\theta^{(l)}$ are interpreted as an implicit M-step.

\paragraph{Transformers and belief propagation.}
Message passing on graphs has long been a template for neural computation.
Neural message passing \cite{gilmer2017messagepassing} unified many graph architectures under one framework, and Factor Graph Neural Networks \cite{zhang2023factorgraph} extend this to learned message passing on factor graphs.
Closer to attention, \cite{zhou2021markovattention} introduce graphical models with attention to capture context-specific independence, applying them to perceptual grouping.
Several recent papers connect transformers directly to BP.
Other work uses BP as a design principle: \cite{lee2025saobp} refine self-attention with a one-step BP update to mitigate attention localization in small-scale models, and \cite{lau2025differentialsyndrome} propose a
differential-attention message-passing transformer.
These papers either establish an equivalence under specific architectural assumptions or use BP to improve the architecture.
The connection between attention and graph computation has also been studied through graph neural networks and graph attention mechanisms~\cite{velickovic2018graph}.
These methods treat attention coefficients as learned edge weights and thereby resemble message passing on graphs.

The present paper differs in its use of BP as the organizing principle: the messages $m_{i,j}^{(t)}(x_j)$, beliefs $\beta_i^{(t)}(x_i)$, and structural indicators $\sigma_{j,i}$ are mapped directly to the components of self-attention.

\paragraph{Dynamic Parameters and Meta-Model Architectures}
Many successful neural architectures can be interpreted as using input-dependent computation.
Hypernetworks \cite{ha2016hypernetworks} established the principle that one network (the meta-parameters) can generate the weights of another.
Mixture-of-Experts models route tokens to different experts using learned gates~\cite{shazeer2017outrageously}, while gated convolutional and recurrent models modulate information flow through input-conditioned gates~\cite{dauphin2017language,hochreiter1997long}.
More recently, selective state-space models such as Mamba use input-dependent state-space parameters to obtain efficient sequence modeling with dynamic behavior~\cite{gu2023mamba}.

These architectures are usually studied as separate design families: routing models, gated models, recurrent models, or state-space models.
The meta-model view proposed here treats them as instances of a broader principle suggested by EM.

\section{Method}
\label{sec:method}
LBP and EM are two classic algorithms specialized at doing inference in different situations.
LBP make inferences on partially observed models with known structure and parameters, while EM on partially observed models with known structure but unknown parameters.
However, with real-life data, not only are they often partially observed but also the underlying structure and parameters are both unknown.
Applying either LPB or EM to real-life data will not work effectively.
We show that the architecture of Transformers combines LPB and EM to make inferences on such complex data.
This combination is what makes Transformers effective.

\subsection{The Underlying Model}
\label{subsec:method-model}
\begin{figure}[htbp]
    \centering
    \includegraphics[width=\textwidth]{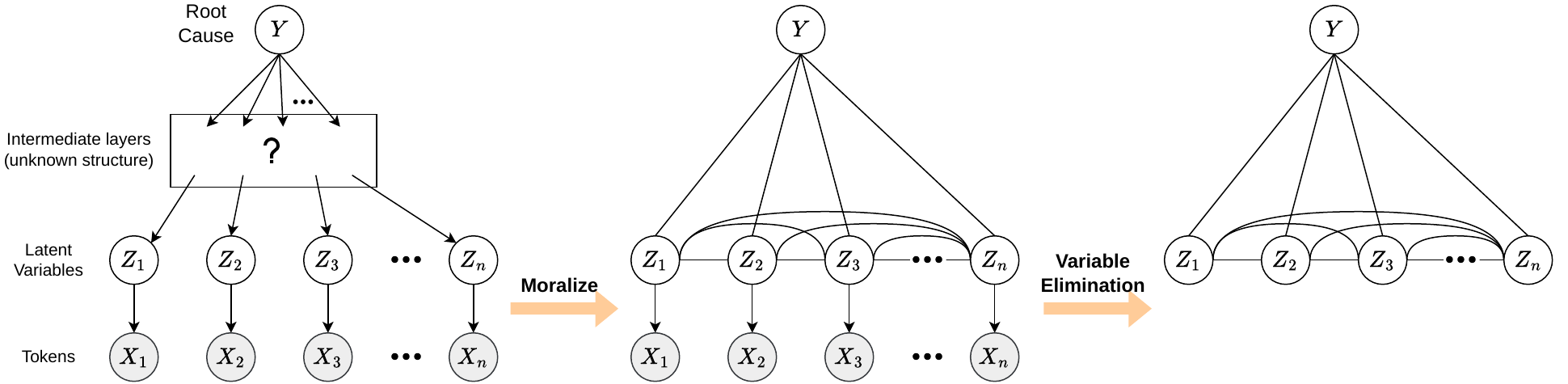}
    \caption{
        The PGM that underlies a encoder Transformer.
        1) On the left is the original BN: tokens $X$ are generated by a sequence of latent variables $Z$.
        $Z$ are further connected to an unknown hierarchy of intermediate layers $S$, which further traces back to a root cause $Y$.
        2) In the middle is the model after moralization: 
        we abstract away $S$ by moralizing the BN over $S$ into a chain graph.
        3) On the right is the model after variable elimination:
        we perform variable elimination on observed tokens $X$.
        The final model is a fully connected MRF on $Z$ and $Y$.
    }
    \label{fig:Transformer_PGM}
\end{figure}
Let us consider a encoder Transformer and examine the probabilistic graphic model (PGM) that underlies it.
Our assumption is that the underlying PGM is a Bayesian Network (BN) as shown on the left of Fig.~\ref{fig:Transformer_PGM}.
The sequence of observed tokens $X=(X_1,\dots,X_n)$ is generated by a sequence of latent variables $Z=(Z_1,\dots,Z_n)$.
$Z$ is further connected to an unknown hierarchical of intermediate layers $S$, which finally trace back to a latent root cause $Y$.
The full factorization is 
\begin{equation}
    p(x,y,z,s) = p(y)p(s|y)\prod_i p(x_i|z_i)p(z_i|s).
\end{equation}
The structure of $S$ is unknown and dynamic, as it can vary from instance to instance.
However, we can abstract it away by moralizing the BN over $S$ into a chain graph, as shown in the middle of Fig.~\ref{fig:Transformer_PGM}.
As a last step, we can perform Variable Elimination on observed leaf variables $X$, as shown to the right of Fig.~\ref{fig:Transformer_PGM}.
In this way, we are able to bake the information of an observation $X_i$ into its latent variable $Z_i$.
This can be seen as imposing a pseudo prior on $Z_i$:
\begin{equation}
    \phi_i(z_i) = p(x_i|z_i),
\end{equation}
where $x_i$ is observed.
As we shall see later, the embedding $e_i$ of a token $x_i$ can be seen as the maximizer of this pseudo-prior:
\begin{equation}
    e_i = \arg\max \phi_i(z_i).
\end{equation}
After the above operations, we are left with a fully connected MRF on $Z$ and $Y$.
We can unify our notation and treat $Y$ as a special latent variable 
\begin{equation}
    Z_{n+1} \doteq Y.
\end{equation}
Let the set of resultant factors on this MRF be $\{\Phi_{i}\mid i=1,\dots,m\}$ for some $m$.
We make the following assumption:
\begin{assumption}
Any factor $\Phi_{i}$ whose arity is larger than 2 can be reasonably approximated by the product of a finite number of pairwise factors.
\end{assumption}
With this assumption, $\{\Phi_{i}\mid i=1,\dots,m\}$ can be approximated by a set of pairwise factors $\{\phi_{i,j}(z_i,z_j)\mid i,j=1,\dots,n+1\}$.
We posit that this MRF with pairwise factors is the model that Transformer works on implicitly.
\begin{assumption}
Encoder Transformers implicitly make inferences on MRFs with pairwise factors that vary from instance to instance.
\end{assumption}

\subsection{Attention as Belief Propagation}
We show that the attention mechanism can be mapped to the max-product BP algorithm.
Given the MRF above, with pairwise factors $\{\phi_{i,j}(z_i,z_j)\mid i,j=1,\dots,n+1\}$, where we define $\phi_{i,i}(z_i,z_i) = \phi_i(z_i)$ as its node potential.
In BP, at time step $t$, every variable $Z_j$ passes a (max-product) message $m_{j,i}^{(t)}$ to every other variable $Z_i$
\begin{equation}
    m_{j,i}^{(t)}(z_i) = \max_{z_j} \left[\phi_{j,i}(z_j,z_i)\prod_{k\in \text{Nb}(j)\backslash \{i\}}m_{k,j}^{(t-1)}(z_j)\right],
\end{equation}
where $\text{Nb}(j)$ is the set of neighbors of node $j$.
Then, at the end of iteration, every variable updates its own belief 
\begin{equation}
    \label{eq:belief}
    \beta_i^{(t)}(z_i) = \prod_{j\in \text{Nb}(i)}m_{j,i}^{(t)}(z_i).
\end{equation}
Note that for the sake of uniformity, here $\text{Nb}(i)$ includes $i$ itself and that the self-message $m_{i,i}(z_i)\doteq\phi_i(z_i)$.
Taking log from both sides, 
\begin{align}
    \label{eq:belief_approx}
    \log\beta_i^{(t)}(z_i) &= \sum_{j\in \text{Nb}(i)} \log m_{j,i}^{(t)}(z_i)\\
                     &= \sum_{j=1}^{n+1} \sigma_{j,i}\log m_{j,i}^{(t)}(z_i),
\end{align}
where $\sigma_{j,i}\in\{0,1\}$ indicates whether nodes $j$ and $i$ are neighbors.
In Transformers, each node (or token position) stores the argument $\dot z_i^{(t)}$ that maximizes its log belief:
\begin{equation}
    \label{eq:maximizer}
    \dot z_i^{(t)} = \arg\max\sum_{j=1}^{n+1} \sigma_{j,i}\log m_{j,i}^{(t)}(z_i).
\end{equation}
If we expand each log message $\log m_{j,i}^{(t)}(z_i)$ at its maximizer $\dot z_{j,i}^{(t)}$ using second-order Taylor expansion, then we can write the belief maximizer as the precision-weighted average of the individual message maximizers:
\begin{equation}
    \dot z_i^{(t)} = \frac{\sum_{j=1}^{n+1} \sigma_{j,i}H_{j,i}\dot z_{j,i}^{(t)}}{\sum_{j=1}^{n+1} \sigma_{j,i}H_{j,i}}, \label{eq:maximizer_approx}
\end{equation}
where $H_{j,i} = -\nabla^2 \log m_{j,i}^{(t)}(z_i)$ is the negative Hessian.
This equation should look familiar by far.
If we further assume that $H_{j,i}$ is isotropic,
\begin{equation}
    \label{eq:hessian_approx}
    H_{j,i} = \alpha_{j,i} I,
\end{equation}
then,
\begin{equation}
    \dot z_i^{(t)} = \frac{\sum_{j=1}^{n+1} \sigma_{j,i}\alpha_{j,i}\dot z_{j,i}^{(t)}}{\sum_{j=1}^{n+1} \sigma_{j,i}\alpha_{j,i}}.
    \label{eq:maximizer_approx_2}
\end{equation}
Self-attention uses the following equations to compute the terms in Eq.\ref{eq:maximizer_approx_2}:
\begin{align}
    \label{eq:self-attention}
    \sigma_{j,i}\alpha_{j,i} &= \exp{(q_i^T k_j)} = \exp{((W_q \dot z_i^{(t-1)})^T (W_k \dot z_j^{(t-1)}))}\\
    \dot z_{j,i}^{(t)} &= v_j = W_v \dot z_j^{(t-1)},
\end{align}
where $q,k,v$ are feature vectors for query, key, and value, and $W_q,W_k,W_v$ are their weight matrices.
We note here that the initial belief $\beta_i^{(0)}(z_i)=\phi_i(z_i)$, mentioned in the last section.
Thus, the embedding of each token can be seen as the maximizer of the prior.

Note that $\sigma_{j,i}$ is capable of learning and adjusting the strength of the connection between two variables, which is equivalent to learning the structure of the underlying MRF.
Consequently, the repetitive applications of layers of self-attention can be seen as performing LBP to infer the latent variables and inferring the structure of the MRF at the same time.

\subsection{Transformers as Expectation-Maximization}
In the last section, we see how self-attention handles hidden variables and unknown structure.
There is one last loose thread: the unknown parameters.
In this section, we show that a Transformer instantiates itself as a hard EM~\cite{neal1998view} to infer the MAP parameters.

Formally, consider supervised learning.
Let $x$ be the input, $y$ the label, $z$ the unobserved latent variables, and $\theta$ the model parameters.
The goal of the model is
\begin{equation}
    \arg\max_\theta \log p(y|x;\theta).
\end{equation}
We have 
\begin{align}
&\log p(y|x;\theta) = \log p(y,z|x;\theta) - \log p(z|y,x;\theta).
\end{align}
Let $\theta^{(l)}$ and $z^{(l)}$ be, respectively, the parameter and the output of layer $l$ of a Transformer.
Then 
\begin{align}
\log p(y|x;\theta) &= \mathbb{E}_{p(z|y,x;\theta^{(l)})}\log p(y,z|x;\theta) - \mathbb{E}_{p(z|y,x;\theta^{(l)})}\log p(z|y,x;\theta) \\
                   &= Q(\theta|\theta^{(l)}) + H(\theta|\theta^{(l)}),
\end{align}
where we have defined
\begin{align}
    Q(\theta|\theta^{(l)})&=\mathbb{E}_{p(z|y,x;\theta^{(l)})}\log p(y,z|x;\theta),\\
    H(\theta|\theta^{(l)})&=-\mathbb{E}_{p(z|y,x;\theta^{(l)})}\log p(z|y,x;\theta).
\end{align}
With Gibbs inequality,
\begin{align}
    &\log p(y|x;\theta) - \log p(y|x;\theta^{(l)}) \\
    &= Q(\theta|\theta^{(l)}) - Q(\theta^{(l)}|\theta^{(l)}) + H(\theta|\theta^{(l)}) - H(\theta^{(l)}|\theta^{(l)}) \\
    &\ge Q(\theta|\theta^{(l)}) - Q(\theta^{(l)}|\theta^{(l)}),
\end{align}
which implies that improvement in $Q(\theta|\theta^{(l)})$ results in at least as much improvement in the log likelihood.

The forward pass of layer $l$ computes $Q(\theta|\theta^{(l)})$, i.e., the E step.
However, instead of computing the exact posterior $p(z|y,x;\theta^{(l)})$, layer $l$ only runs one iteration of approximate max-product BP to get an estimated maximizer $z^{(l)}$.
Then we have
\begin{equation}
    Q(\theta|\theta^{(l)}) = \log p(y,z^{(l)}|x;\theta).
\end{equation}
The job of the next layer $l+1$ is to maximize $Q(\theta|\theta^{(l)})$, i.e., the M step.
Specifically, The parameter $\theta^{(l+1)}$ of the next layer corresponds to the parameter $\theta$ that maximizes $Q(\theta|\theta^{(l)})$:
\begin{equation}
    \theta^{(l+1)} = \arg\max_\theta \log p(y,z^{(l)}|x;\theta).
\end{equation}

See Fig.~\ref{fig:dnn_as_em} for an illustration.
In summary, we can view a layer of a Transformer as an iteration of the EM.
The forward pass of a layer corresponds to the E step, where the latent variables for each token are updated on the basis of last layer.
The change in the parameters of a layer compared to its previous layer implicitly corresponds to the M step, where the model parameters are updated.

The EM point of view has several consequences.
Here, we list a few we can think of.
\begin{itemize}
    \item It provides key insight into the in-context learning capability of Transformers, to be elaborated in the next section.
    \item Training a model is equivalent to learning an EM algorithm.
    \item It partly explains the efficacy of ResNet\cite{he2016deep}, where each layer is only responsible for learning an increment to latent variables.
    \item The parameters of different layers are essentially different estimates of the same set of parameters.
    This may partly explain why DNNs do not suffer from significant overfitting~\cite{zhang2021understanding}, because the effective number of parameters remains constant as layers grow.
    This constitutes a complementary point of view to some previous ones~\cite{neyshabur2017exploring,poggio2020complexity,belkin2019reconciling}.
\end{itemize}

\subsection{Meta-Models}
An important consequence of the EM point of view lies in the M-step: $\theta^{(l+1)} = \arg\max_\theta \log p(y,z^{(l)}|x;\theta)$.
The M-step says that the parameters $\theta^{(l+1)}$ of layer $l+1$ should be dependent on its input $z^{(l)}$, which implies a procedure that generates parameters based on the input, prior to the forward pass.
This leads to a natural model design paradigm defined by the following two sequential passes:
\begin{equation}
\begin{aligned}
    \theta &= F_\Theta(z),\\
    z' &= f_\theta(z),
\end{aligned}
\end{equation}
where $f$ is the forward pass, $F$ the meta-forward pass, $\theta$ the parameters, $\Theta$ the meta-parameters, and $z$ and $z'$ respectively the input and output.
In the meta-forward pass, $F$ takes $z$ as input and determines what parameters to use in the forward pass.
This dynamic generation of parameters is what makes in-context learning possible.
In fact, many familiar deep architectures are members of meta-models.

\paragraph{Self-attention as a meta-model}.
Let $z,z'\in\mathbb{R}^{d\times n}$ be the input and output and $Q,K,V\in\mathbb{R}^{d\times d}$ be the query, key, and value matrices.
The following setup is equivalent to self-attention:
\begin{equation}
    \begin{aligned}
        A &\doteq F_{\{Q,K\}}(z) = \sigma((Qz)^T(Kz)/\sqrt d),\\
        z' &\doteq f_A(z) = AzV,
    \end{aligned}
\end{equation}
where $\sigma$ is the Softmax operation.

\paragraph{MoE as a meta-model}.
In the Mixture of Experts (MoE) models, let $K$ be the number of experts, $W\in \mathbb R^{K\times d'\times d}$ the parameters for each expert, $ R\in \mathbb{R}^{K\times d}$ the routing parameters, $z\in\mathbb R^d$ and $z'\in\mathbb R^{d'}$ respectively the input and output.
The following setup is equivalent to MoE:
\begin{equation}
    \begin{aligned}
        w &\doteq F_{R}(z) = \sigma(R z),\\
        z' &\doteq f_{w}(z) = w^T W z,
    \end{aligned}
\end{equation}
where $\sigma$ is the Softmax operation.
Note that MoE is essentially a meta version of the linear layer.

Meta-models also include LSTM~\cite{hochreiter1997long} and Mamba~\cite{gu2023mamba}, which we will leave to interested readers.
Meta-models entail a two-pass procedure of \textit{thinking about your own thinking}, which could possibly be likened to the meta-cognition~\cite{flavell1979metacognition} in psychology.

\paragraph{Meta-Linear model}.
Here, we also propose a ordinary design for the Meta-Linear layer.
Let $f$ be a linear operation with parameter $\theta \in \mathbb R^{d\times d'}$ and $F$ its meta-linear operation with parameter $\Theta \in \mathbb R^{k\times d'\times d}$.
Then 
\begin{equation}
    \begin{aligned}
        \theta &= F_\Theta(z) = \Theta^T \sigma(z),\\
        z' &= f_\theta(z) = \theta^T z,
    \end{aligned}
\end{equation}
where $z \in \mathbb R^{d}$ and $z' \in \mathbb R^{d'}$.
$\sigma$ is some (parameterized) nonlinearity that projects $d$-dimension to $k$-dimension, if $k\neq d$.
If we let $\sigma(x) = \text{Softmax}(x^T R)$, where $R$ is $d$-by-$k$, and interpret $k$ as the number of experts, then this is equivalent to a $k$-expert MoE model.
The parameter count (excluding nonlinearity) of a Meta-Linear layer is $k$ times that of a linear layer.
See Fig.~\ref{fig:meta-models} for an illustration.

\subsection{Generalized Meta-Models}
We can further generalize the idea of meta-models to higher orders.
Here, we take the linear layer as an example.
For an $n$-th order meta-linear (or meta$^n$-linear) layer with input $z\in \mathbb{R}^d$ and output $z'\in \mathbb{R}^{d'}$, the procedure becomes
\begin{gather}
        \Theta^{(n-1)} = F^{(n)}(z) = \Theta^{(n)} \sigma(z),\\
        \vdots\notag\\
        \Theta^{(i-1)} = F^{(i)}(z) = \Theta^{(i)} \sigma(z),\\
        \vdots\notag\\
        \theta=\Theta^{(0)} = F^{(1)}(z) = \Theta^{(1)} \sigma(z),\\
        z'=\Theta^{(-1)} = F^{(0)}(z) = \Theta^{(0)} z,
\end{gather}
where $\Theta^{(i)} \in \mathbb R^{d'\times d\times k_1\dots\times k_i}$ is called meta$^i$-parameter, and $F^{(i)}$ the meta$^i$-forward pass.
Note, especially, that $\Theta^{(-1)} = z'$ is the output, $\Theta^{(0)} = \theta$ and $F^{(0)} = f$.
The parameter count of a generalized meta-linear layer of order $n$ is
\begin{equation}
    \text{\#parameter} = d'dk^{n},
\end{equation}
which grows exponentially with its order.

\section{Experiments}

\subsection{Retrieving the Probabilistic Graphs}
If Transformers are indeed making inferences in the PGMs mentioned in Section~\ref{subsec:method-model}, we can somehow retrieve the PGMs from their outputs.
In an attempt to do this, we do the following experiments.
Given an input text, we take the output feature vector of each token from the last layer of a Transformer, and remove vectors of those non-semantic or special tokens, e.g. punctuations, newlines, </s>, and <pad>.
However, the class tokens, e.g. <s> and [CLS], used for final classification, if present, are retained.
We compute the similarity matrix $\mathcal S$, where each entry is the cosine similarity between two feature vectors.
\begin{equation}
    \mathcal{S}_{i,j} = \text{cosine\_similarity}(z_i, z_j).
\end{equation}
Then we run single-linkage agglomerative hierarchical clustering algorithm on the matrix to detect hierarchical communities and plot the results as dendrograms.
We do experiments on multiple movie reviews and real-life texts from X (formerly Twitter) with multiple Transformers from Huggingface.
Models include:
\begin{itemize}
    \item twitter-roberta-base-sentiment-latest~\cite{loureiro2022timelms}. A RoBERTa-base model pretrained on about 124M tweets (2018-2021) and fine-tuned for three-class sentiment classification (negative, neutral, positive) on the TweetEval benchmark (Loureiro et al., 2022; Barbieri et al., 2020).
    Its class token is <s>.
    \item distilbert-base-uncased~\cite{sanh2019distilbert}. A compact (66M parameters), general-purpose English encoder obtained by distilling BERT-base (Sanh et al., 2019).
    Its class token is [CLS].
    \item BGE-M3~\cite{chen2024bge}. A multilingual embedding model built on XLM-RoBERTa (about 568M parameters).
    Its class token is <s>.
    \item Qwen3-Embedding-0.6B~\cite{zhang2025qwen3}. A 0.6B-parameter text embedding model derived from the Qwen3 foundation model.
    It does not have a class token.
\end{itemize}

\begin{figure}[htbp]
  \centering
  \setlength{\tabcolsep}{1pt}      
  \renewcommand{\arraystretch}{1.1}
  \begin{tabular}{>{\raggedright\arraybackslash}b{2.5cm} *{4}{c}}       
     & \small{Roberta} & \small{bge-m3} & \small{distilbert} & \small{Qwen3-Embedding} \\ [2pt]
    \small{"Toy Story 5 is a beautiful, timely story that captures the moment and examines our relationship with technology and real, human connection."}
      & \includegraphics[width=0.21\linewidth]{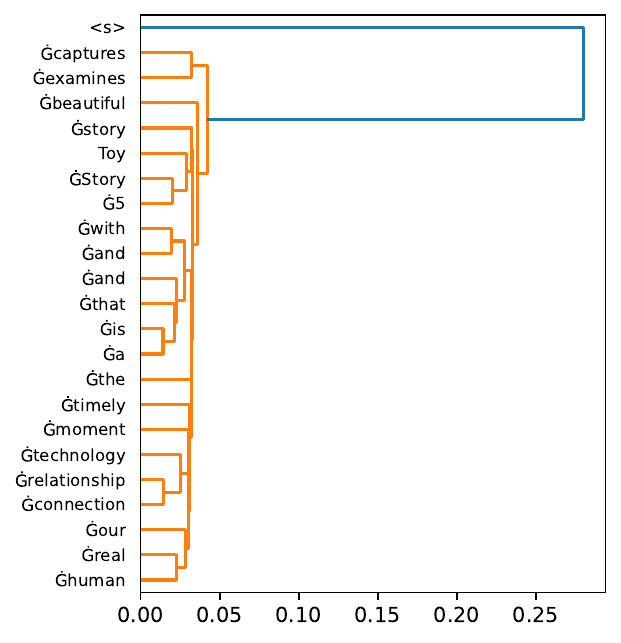}
      & \includegraphics[width=0.21\linewidth]{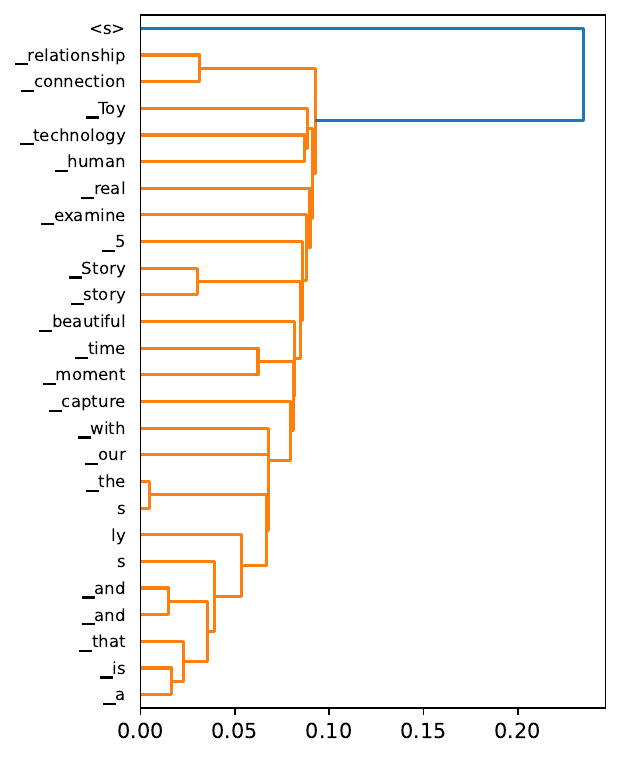}
      & \includegraphics[width=0.21\linewidth]{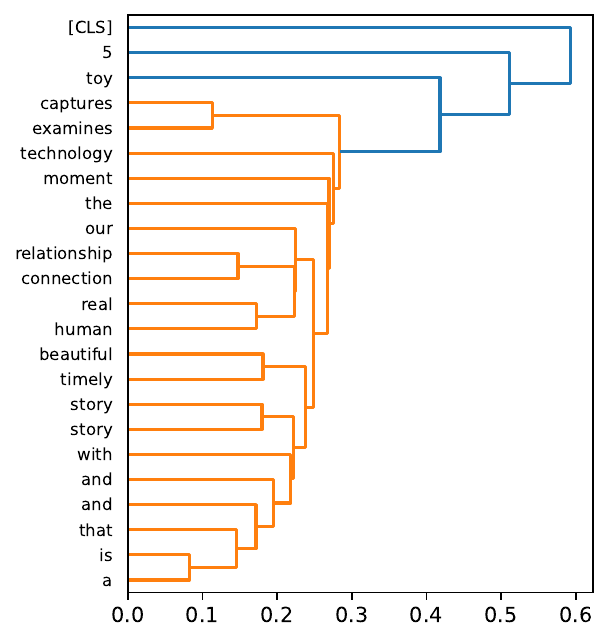}
      & \includegraphics[width=0.21\linewidth]{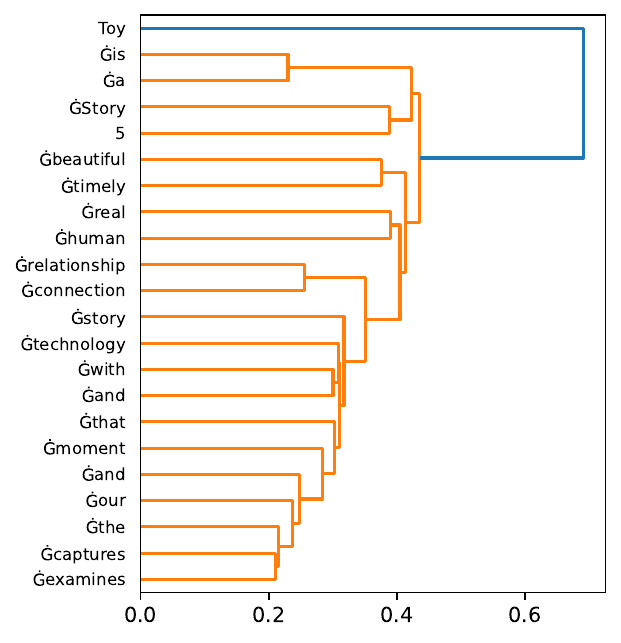}
      \\
    \small{"Remember that I predicted a long time ago that President Obama will attack Iran because of his inability to negotiate properly-not skilled!"}
      & \includegraphics[width=0.21\linewidth]{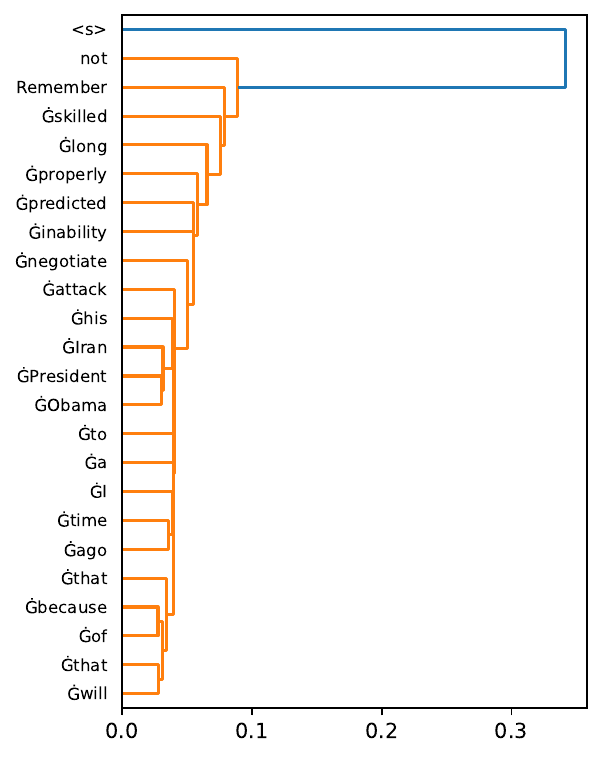}
      & \includegraphics[width=0.21\linewidth]{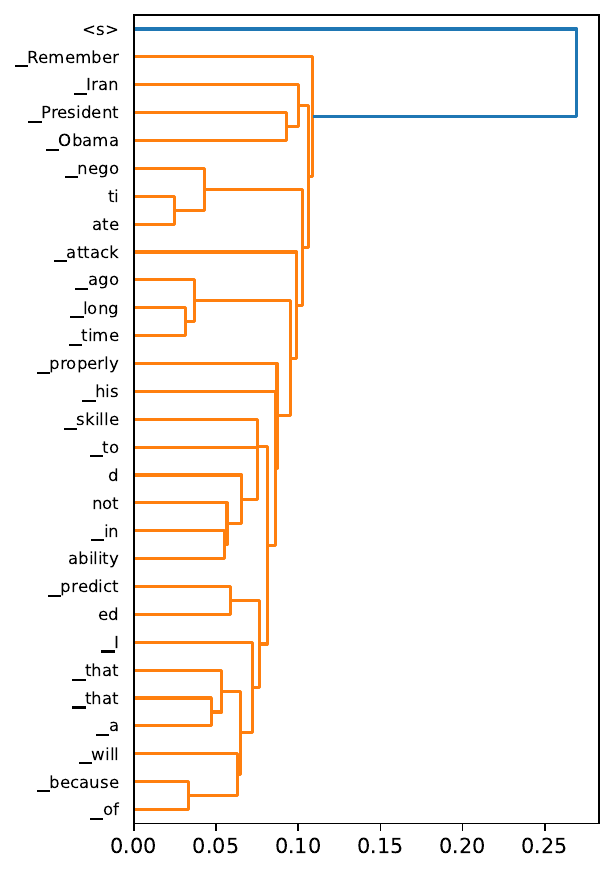}
      & \includegraphics[width=0.21\linewidth]{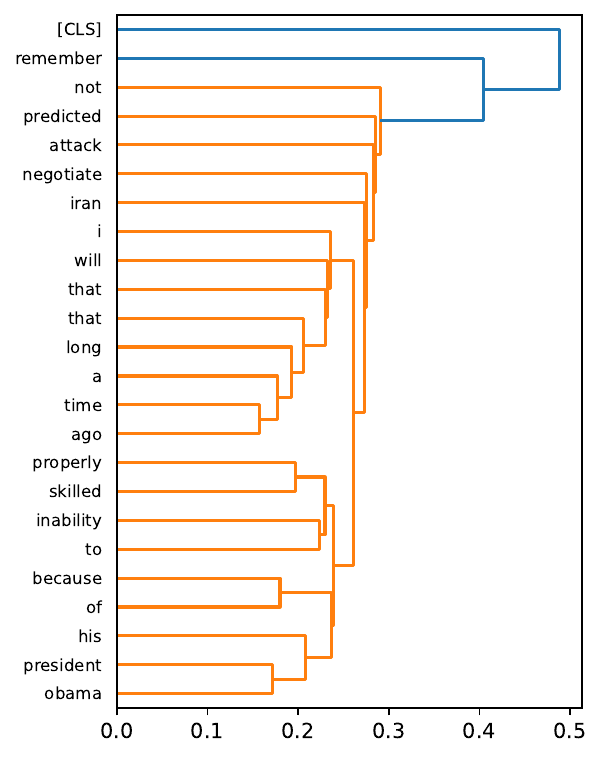}
      & \includegraphics[width=0.21\linewidth]{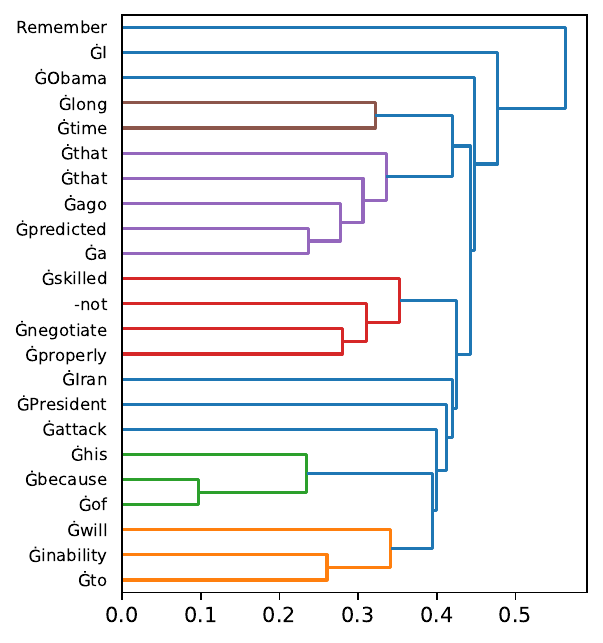}
      \\
    \small{"I paid for TwitterBlue to support Elon and the platform. He cleaned up the place but lost a ton of advertising revenue. I'm sure there are many like-minded users who cared less about getting their bird painted blue and continue paying their monthly dues."}
      & \includegraphics[width=0.21\linewidth]{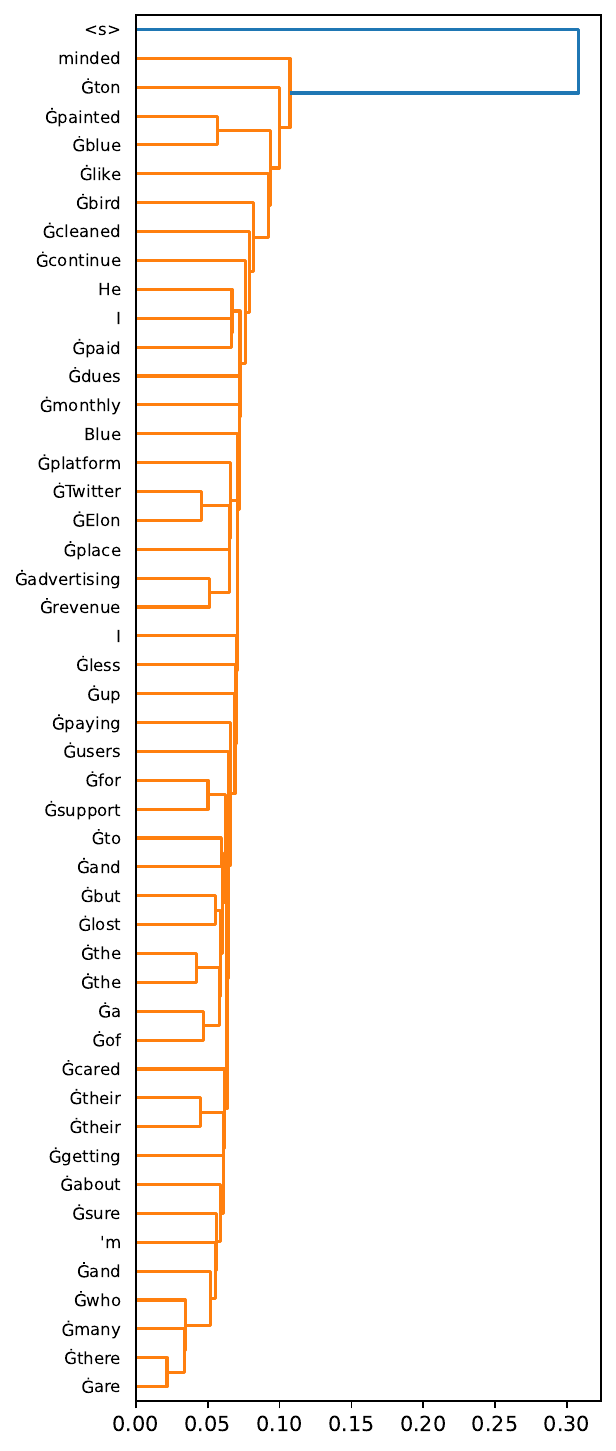}
      & \includegraphics[width=0.21\linewidth]{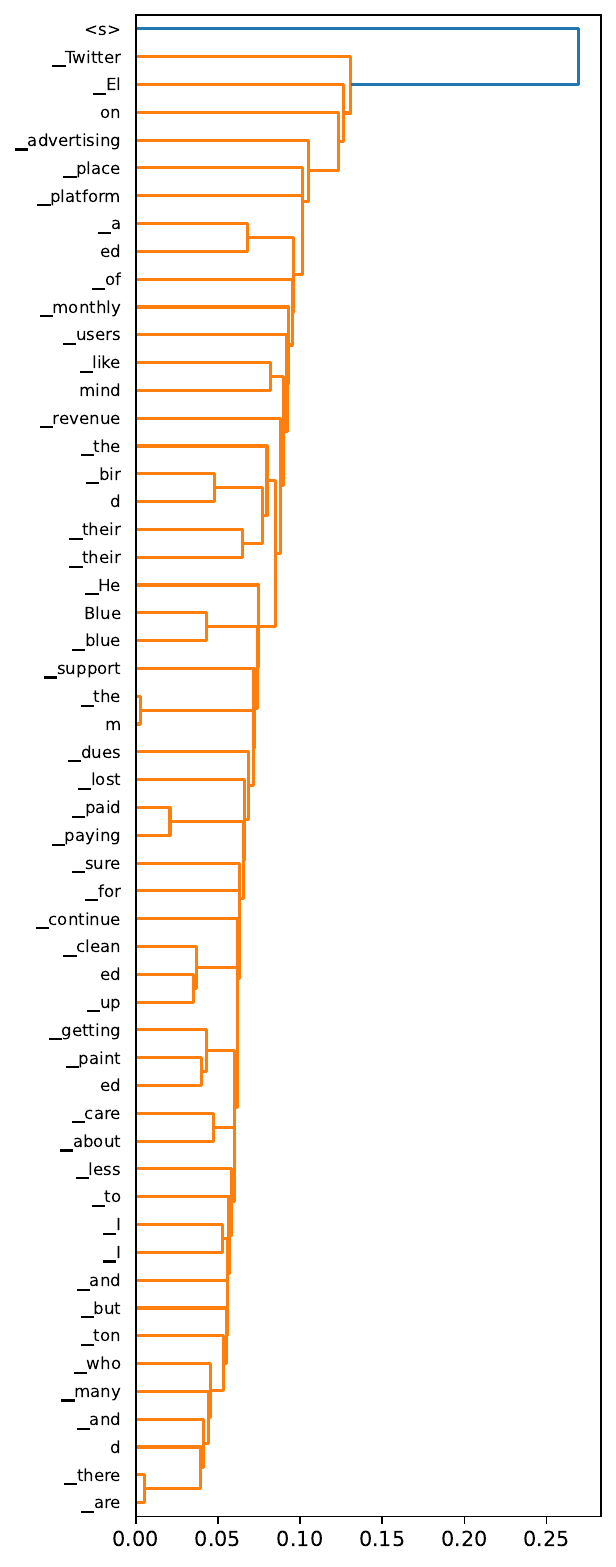}
      & \includegraphics[width=0.21\linewidth]{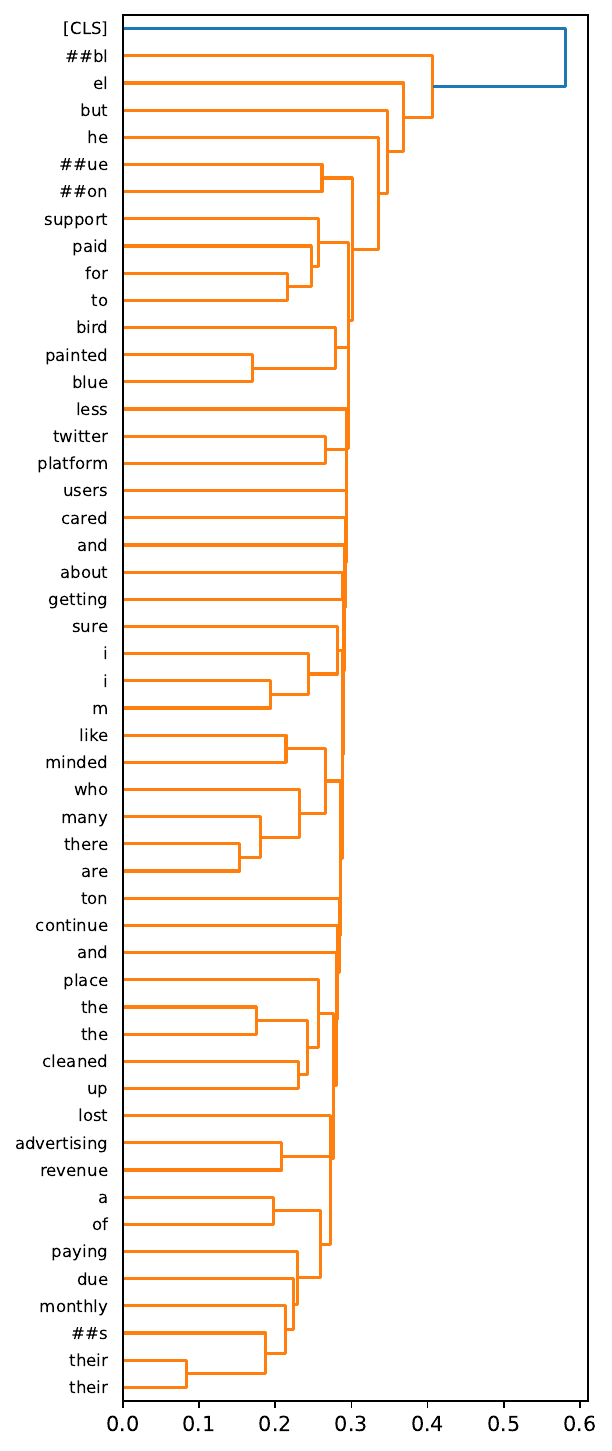}
      & \includegraphics[width=0.21\linewidth]{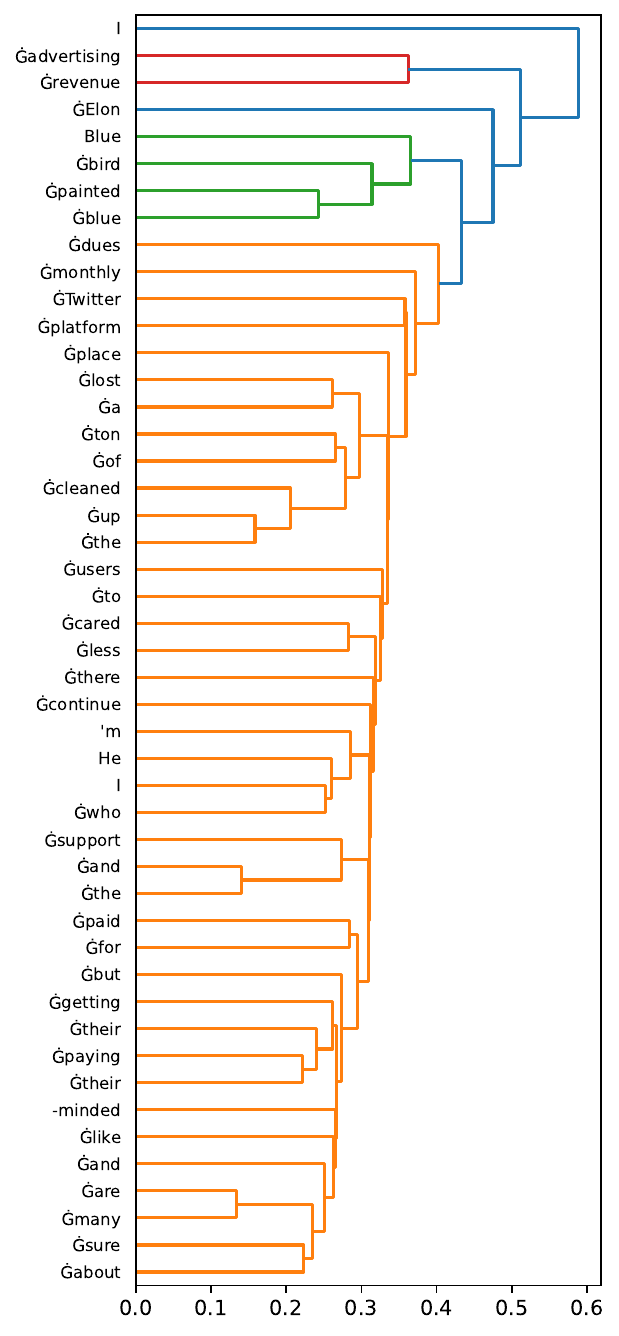}
  \end{tabular}
  \caption{The agglomerative hierarchical clustering dendrograms for various models and texts. We run single-linkage agglomerative hierarchical clustering algorithm on the features of Transformers' last layer to detect hierarchical communities and plot the results as dendrograms. We can see that the class tokens are lastly merged into the hierarchy, supporting our theory.}
  \label{fig:retrieve-pgm-exp-result}
\end{figure}

The results are shown in Fig.~\ref{fig:retrieve-pgm-exp-result}.
A few important observations can be made from the results.
\begin{itemize}
    \item Most compound words, phrases, and collocations, e.g. "like-minded", "cleaned up", are immediately grouped together by Transformers.
    This indicates Transformers are aware of the fact that they form a single entity and each word in a compound is generated by that entity.
    \item These recognized entities are further merged with other related tokens or entities, forming a hierarchy.
    \item Some pronouns are immediate neighbors to some other entities to which they refer in the original texts.
    \item The class token is significantly distant from other communities and is lastly merged into the hierarchy, which indicates that Transformers recognize it as the root cause.
\end{itemize}


\subsection{Meta-Linear models}
We evaluate the expressiveness and classification capabilities of Linear and Meta-Linear models on synthetic data generated from Gaussian Mixture Models (GMMs).

\begin{figure}[htbp]
    \centering
    \begin{subfigure}[b]{0.48\textwidth}
        \centering
        \includegraphics[width=\textwidth]{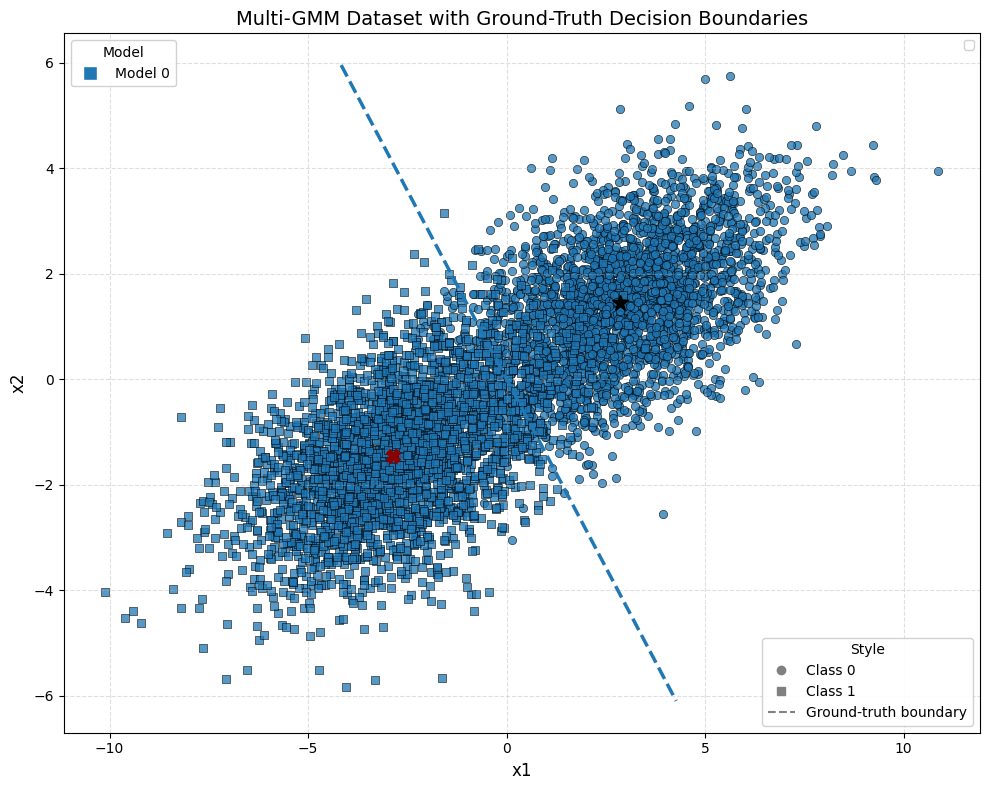}
        \caption{Number of GMMs $M=1$}
        \label{fig:gmm_synthetic_data-11}
    \end{subfigure}
    \hfill
    \begin{subfigure}[b]{0.48\textwidth}
        \centering
        \includegraphics[width=\textwidth]{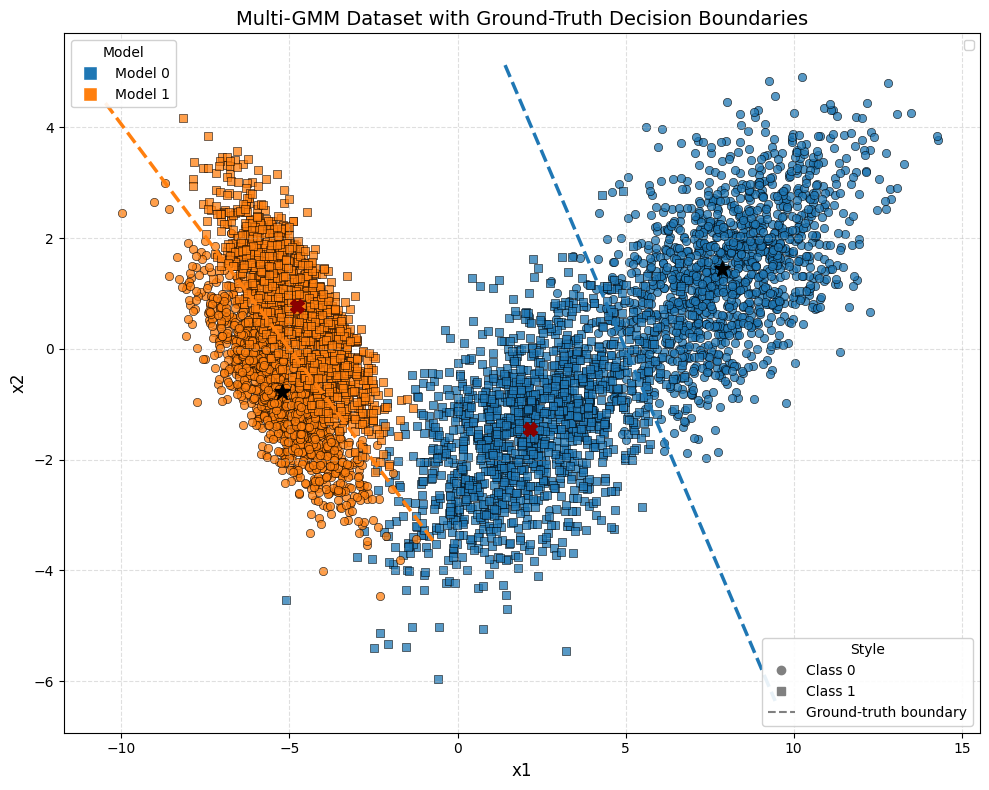}
        \caption{$M=2$}
        \label{fig:gmm_synthetic_data-12}
    \end{subfigure}
    \vspace{0.1cm} 
    \begin{subfigure}[b]{0.48\textwidth}
        \centering
        \includegraphics[width=\textwidth]{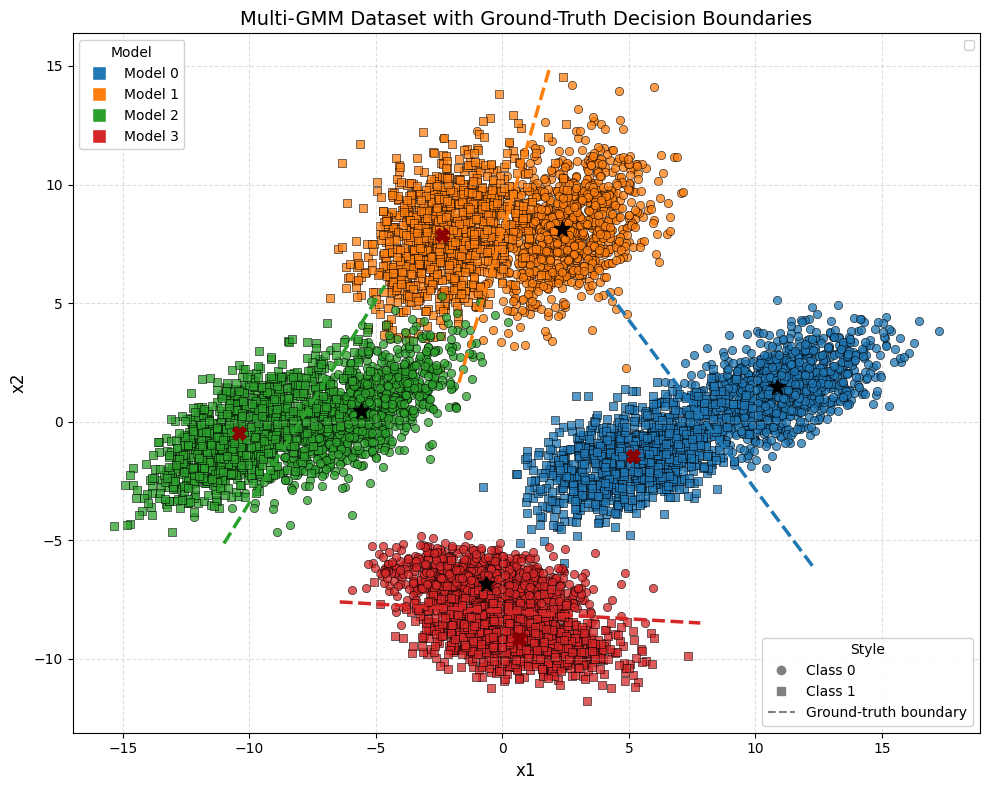}
        \caption{$M=4$}
        \label{fig:gmm_synthetic_data-21}
    \end{subfigure}
    \hfill
    \begin{subfigure}[b]{0.48\textwidth}
        \centering
        \includegraphics[width=\textwidth]{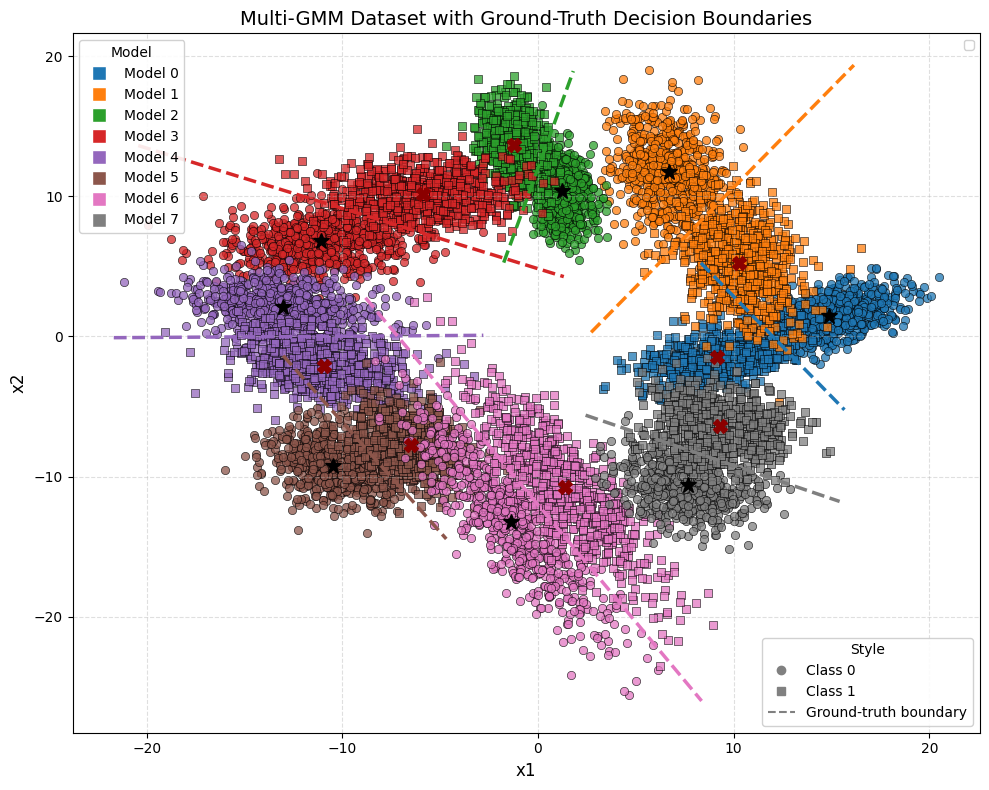}
        \caption{$M=8$}
        \label{fig:gmm_synthetic_data-22}
    \end{subfigure}
    \caption{The generated synthetic multi-GMM data sets consisting of different number of GMM models $M=1,2,4,8$.} with ground-truth Bayes decision boundaries.
    \label{fig:gmm_synthetic_data}
\end{figure}

To evaluate model performance across multiple sub-distributions, we generated $M$ distinct 2D datasets ($d=2$), each consisting of two Gaussian mixture distributions representing Class 0 ($\mathcal{N}_1$) and Class 1 ($\mathcal{N}_2$).
We repeat the experiments for various number of GMM models of $M=1,2,4,8$.
The generated data are visualized in Fig.\ref{fig:gmm_synthetic_data}.

\textbf{Covariance Sampling}: For each dataset $c \in \{0, \dots, M-1\}$, a shared covariance matrix $\mathbf{\Sigma}_c$ is sampled from a Wishart distribution ($\text{df} = d + 3 = 5$) with scale matrix $\mathbf{S} = 0.8 \mathbf{I}_2$.

\textbf{Class Separation}: The difference in class means $\boldsymbol{\delta} = \boldsymbol{\mu}_1 - \boldsymbol{\mu}_0$ is scaled according to a target Mahalanobis distance $d_M = 3.2$, yielding high linear separability (~93–98\%) per sub-model.

\textbf{Data Splitting}: Each class contains $N = 4000, $ samples per dataset.
The combined dataset is split into training (75\%), evaluation (15\%), and test (15\%) using stratified sampling.

\textbf{Theoretical classification accuracy}: We also compute the theoretical upper bound for classification accuracy on test set using the optimal Bayes decision boundary from training set.
For each sub-model $c$, the decision boundary is given by the linear discriminant score $f(\mathbf{x}) = \mathbf{w}_c^T \mathbf{x} + b_c$, where
  $$\mathbf{w}_c = \mathbf{\Sigma}_c^{-1}(\boldsymbol{\mu}_1 - \boldsymbol{\mu}_0)$$$$b_c = -\frac{1}{2}(\boldsymbol{\mu}_1 + \boldsymbol{\mu}_0)^T \mathbf{w}_c$$
Predictions are assigned based on $\text{sign}(f(\mathbf{x}))$, establishing the ground-truth classification accuracy on the test split.

\textbf{Models compared}: We compare several custom neural network layer formulations designed to dynamically weight input features:
\begin{enumerate}
\item Linear: A vanilla linear layer of the form $f(\mathbf{x}) = W\mathbf{x} + b$.
\item Meta-Linear: A implementation evaluating $f(\mathbf{x}) = \sigma(\mathbf{x}^T) (W \mathbf{x} + B)$, where meta-weight $W$ is of shape $(k, d', d)$ and meta-bias $B$ of $(k,d')$.
We compare different choices of non-linearity $\sigma$:
    \begin{itemize}
        \item Identity $\sigma(x)=x$
        \item $\text{Swish}(x)=x\odot \text{Sigmoid}(x)$
        \item Parametrized $\text{Swish}(x)=x\odot \text{Sigmoid}(Vx)$
        \item $\text{SwiGLU}(x) = Vx \odot \text{Sigmoid}(Vx) \odot Wx$
        \item Ground-truth MoE: We can make Meta-Linear equivalent to MoE by $\sigma(x)=\text{Softmax}(Rx)$.
        We let number of experts be the ground-truth $k=M$.
    \end{itemize}
\end{enumerate}

\begin{figure}[htbp]
    \centering
    \begin{subfigure}[b]{\textwidth}
        \centering
        \includegraphics[width=0.8\textwidth, keepaspectratio]{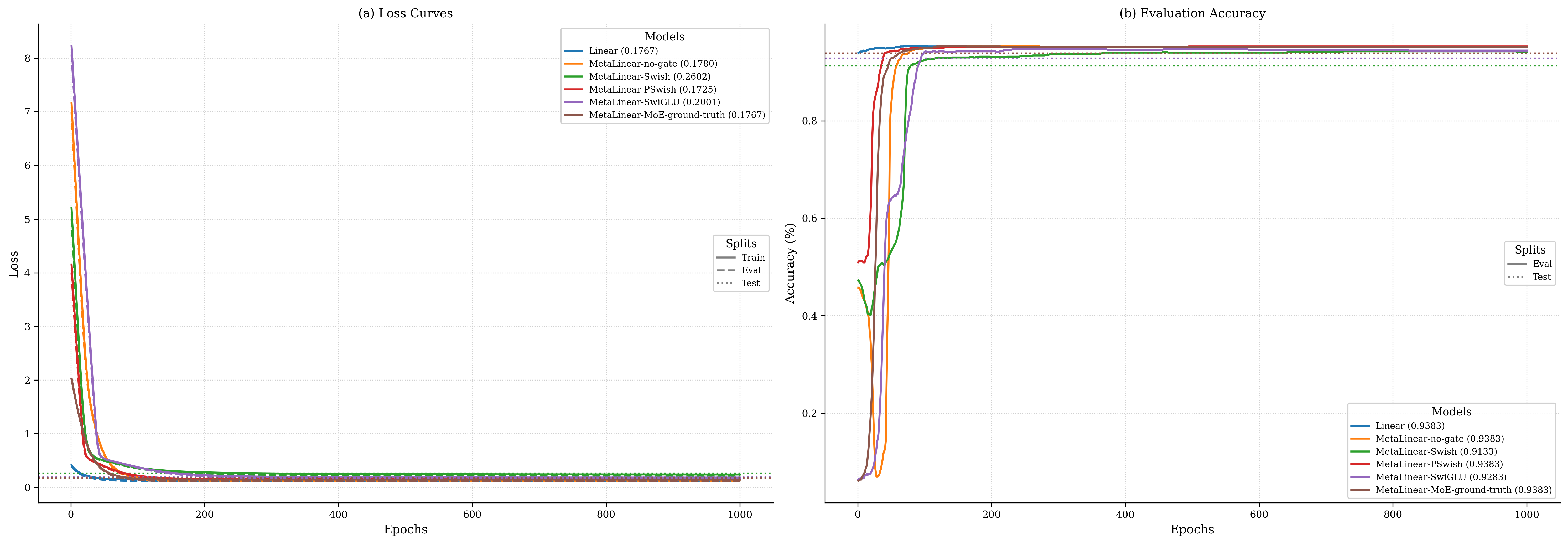}
        \caption{$M=1$}
        \label{fig:meta-models-exp-result-1}
    \end{subfigure}
    \vspace{0.1cm} 
    \begin{subfigure}[b]{\textwidth}
        \centering
        \includegraphics[width=0.8\textwidth, keepaspectratio]{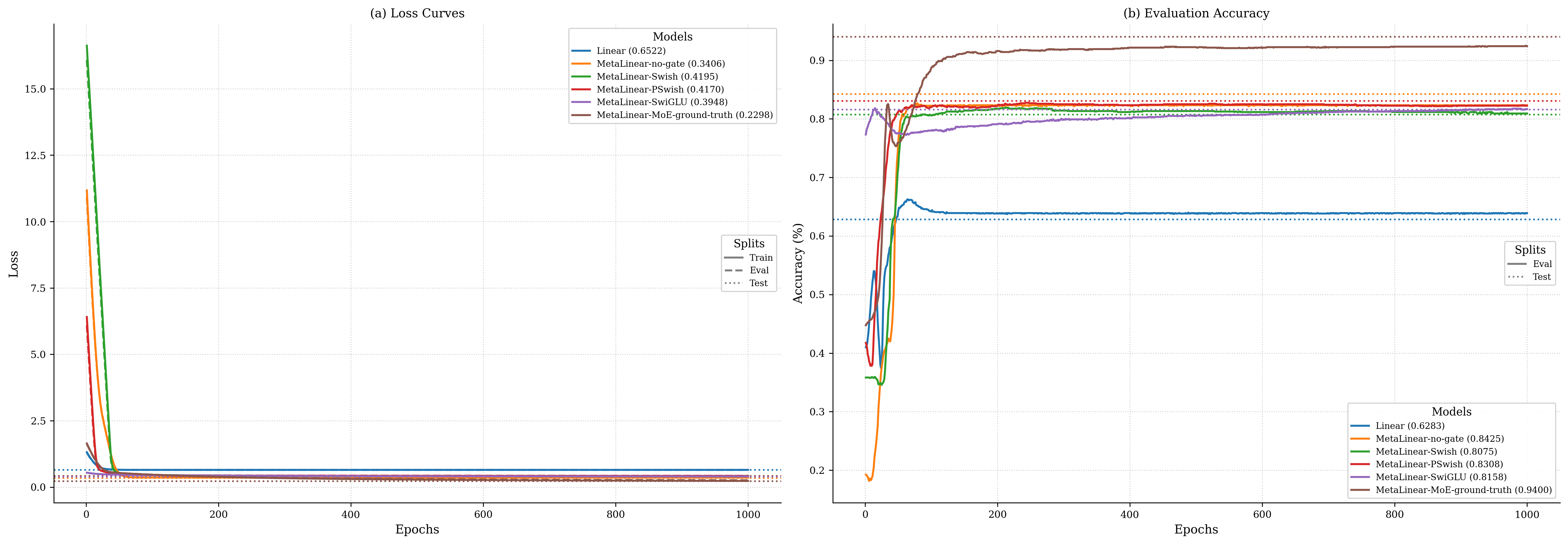}
        \caption{$M=2$}
        \label{fig:meta-models-exp-result-2}
    \end{subfigure}
    \vspace{0.1cm}
    \begin{subfigure}[b]{\textwidth}
        \centering
        \includegraphics[width=0.8\textwidth, keepaspectratio]{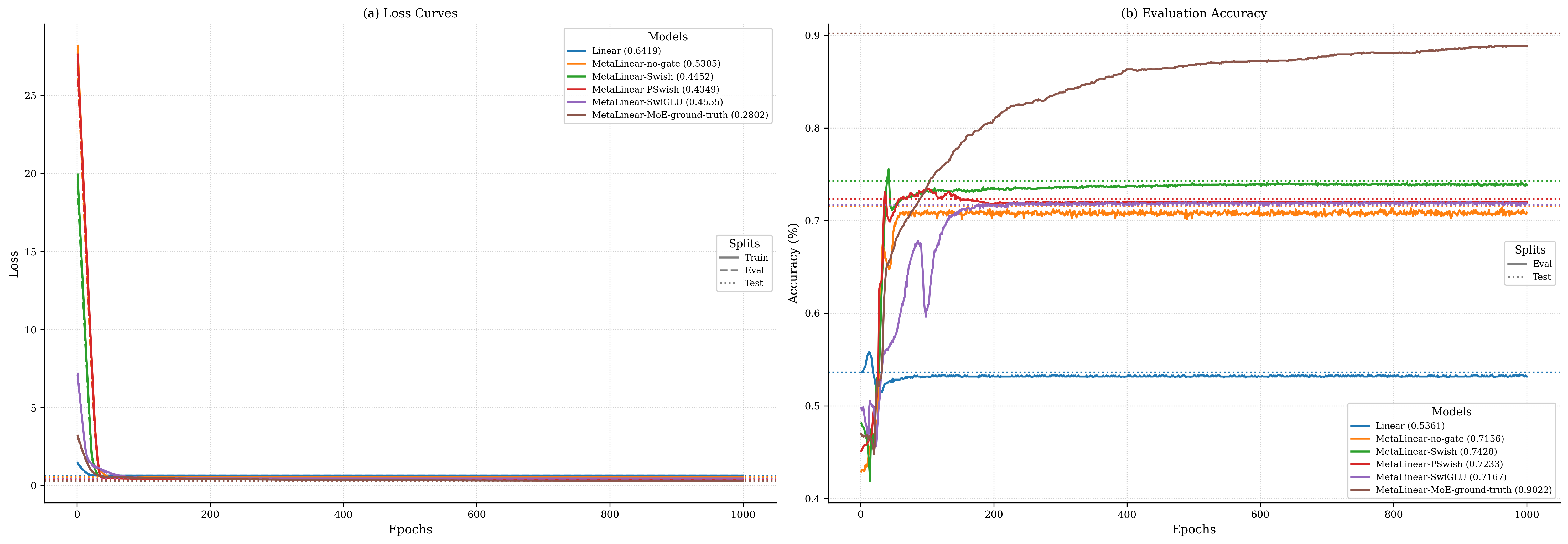}
        \caption{$M=4$}
        \label{fig:meta-models-exp-result-3}
    \end{subfigure}
    \vspace{0.1cm}
    \begin{subfigure}[b]{\textwidth}
        \centering
        \includegraphics[width=0.8\textwidth, keepaspectratio]{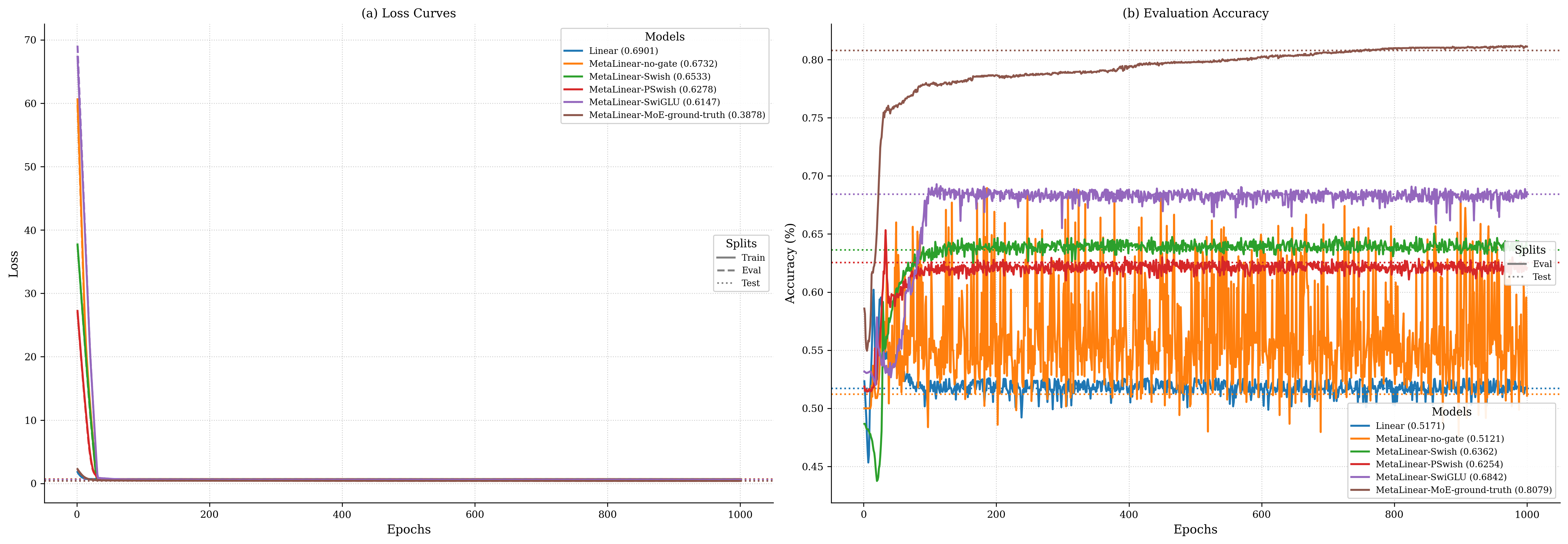}
        \caption{$M=8$}
        \label{fig:meta-models-exp-result-4}
    \end{subfigure}
    \caption{Training results for comparing different linear models for different $M$.}
    \label{fig:meta-models-exp-result}
\end{figure}

We use Binary Cross-Entropy optimized via ADAM~\cite{kingma2014adam} with a learning rate $\eta = 0.0001$ and batch size 32 for 1000 epochs.
Loss and accuracy are monitored at each epoch on the training and evaluation set, with final generalization accuracy evaluated on the held-out test dataset.
The results are shown in Fig.~\ref{fig:meta-models-exp-result}.

Based on the experimental evaluation, the following conclusions can be drawn:
\begin{itemize}
    \item Baseline Degradation with Multi-Modal Complexity: The vanilla Linear layer is highly effective (accuracy 0.9383) only for simple, linearly separable data ($M=1$).
    As $M$ increases to $M=2, 4,$ and $8$, the classification task becomes significantly non-linear, causing the vanilla Linear model's accuracy to degrade to 0.6283, 0.5361, and ultimately 0.5171, respectively.
    \item Superior Expressiveness of Meta-Linear Models: Dense Meta-Linear formulations successfully overcome the limitations of the baseline by dynamically weighting weight matrix.
    Across all multi-modal datasets ($M > 1$), the Meta-Linear variants consistently outperform the standard Linear layer, demonstrating a greater capacity to model complex decision boundaries.
    \item Efficacy of Non-Linear Gating Functions: The specific choice of non-linearity ($\sigma$) plays a critical role in the model's classification capability.
    Advanced gating mechanisms, particularly SwiGLU and Parametrized Swish (PSwish), provide better adaptability than simpler formulations like the identity (no-gate) function.
    For instance, at the highest complexity ($M=8$), the Meta-Linear SwiGLU variant achieves an evaluation accuracy of 0.6842, substantially outperforming the vanilla baseline.
    \item Ground-Truth MoE as the Performance Upper Bound: The Ground-truth MoE, which explicitly sets the number of experts equal to the number of sub-models ($k=M$), establishes the theoretical upper limit for classification accuracy in these experiments.
    It remains highly robust against dataset complexity, maintaining top-tier accuracies of 0.9400 at $M=2$, 0.9022 at $M=4$, and 0.8079 at $M=8$.
\end{itemize}

\section{Conclusion}
This paper has developed a probabilistic-inference interpretation of Transformers as Loopy Belief Propagated Expectation-Maximization.
The central problem is to explain how self-attention, depth, latent representation updates, graph-structure selection, and layerwise parameter variation can be understood within one theoretical framework.
We argue that self-attention approximates LBP in a dynamically structured MRF, where attention weights $\sigma_{j,i}$ encode learned structural affinities and value vectors approximate message-specific maximizers.
We further interpreted successive Transformer layers as an EM-like process in which the forward pass updates latent variables and the transition from $\theta^{(l)}$ to $\theta^{(l+1)}$ acts as an implicit M-step.

This perspective suggests that several known properties of deep architectures have a common origin.
Residual depth can be understood as iterative refinement of latent variables, while dynamic computation mechanisms such as self-attention, MoE, gated units, LSTM, and selective state-space models can be viewed as forms of input-conditioned parameter generation.
Empirical findings are strongly supportive of our claims.
Even without empirical claims, the derivations provide a unified language via graphical-model views of Transformer computation.

Future work should make this interpretation quantitatively testable.
First, one can measure whether attention maps behave like inferred MRF structures by comparing them with known dependency graphs in synthetic graphical-model tasks.
Second, one can test the EM interpretation by tracking whether layer-wise representations and parameters improve a surrogate $Q(\theta|\theta^{(l)})$ objective across depth.
Third, the meta-model principle motivates new architectures that generate parameters with controlled order and sparsity, allowing a direct study of the tradeoff between expressive dynamic computation and the exponential parameter growth of generalized meta-models.

\bibliographystyle{plain}
\bibliography{references}

\end{document}